\documentclass{article}

     \PassOptionsToPackage{numbers, compress}{natbib}
\usepackage{tabularx}
\usepackage{soul,color,xcolor}
\usepackage{wrapfig}
\usepackage{graphicx}
\usepackage{enumitem}

  \usepackage[dblblindworkshop, final]{neurips_2026}
\workshoptitle{Trustworthy AI for Good}
\usepackage[utf8]{inputenc} % allow utf-8 input
\usepackage[T1]{fontenc}    % use 8-bit T1 fonts
\usepackage{hyperref}       % hyperlinks
\usepackage{url}            % simple URL typesetting
\usepackage{booktabs}       % professional-quality tables
\usepackage{amsfonts}       % blackboard math symbols
\usepackage{nicefrac}       % compact symbols for 1/2, etc.
\usepackage{microtype}      % microtypography
\usepackage{xcolor}         % colors

\title{Whose Ground Truth? \\ Embracing Ambiguity in Human-Centered AI}

\author{
Jingyao Wu$^{1}$\thanks{Corresponding author},
~Mohammad Tariqul Islam$^{1}$,
Per Rådberg Nagbøl$^{2}$,
Julie Gerlings$^{2,3}$,\\
\textbf{Giovanni Leoni$^{2}$,
Georgiana Fryza$^{2}$,
Lars Pilegaard Thomsen$^{2,4}$, Jakob Mainz$^{2,4}$} \\
\\[-0.3em]
$^{1}$Massachusetts Institute of Technology, USA \\
$^{2}$Novo Nordisk A/S, Denmark \\
$^{3}$Copenhagen Business School, Denmark\\
$^{4}$Aalborg University, Denmark\\
\\[-0.3em]
\texttt{\{jingyaow, mhdtariq\}@mit.edu} \\
\texttt{\{PYRL, IJGE, GLEO, GNBV, LPTH, JTXM\}@novonordisk.com}
}

\begin{document}

\maketitle

\begin{abstract}
  %other potential title: Beyond the Duck-Rabbit: Human-Centered AI Should Embrace Ambiguity
 As AI systems increasingly interact with people and make decisions about them, understanding human interpretations becomes an important part of developing human-centered AI. Conventional machine learning and AI systems are largely developed under the assumption that a single definitive ground truth exists, with variability in human annotations often resolved through aggregation or treated as noise. However, for many human-centered tasks, human interpretation is inherently ambiguous, and multiple interpretations of the same input may be simultaneously reasonable and valid. Reducing such ambiguity to a single target risks overlooking meaningful information about the diversity of human perception, judgment, and experience. In this position paper, we call for a shift towards modeling the interpretation space of plausible human judgments, while distinguishing meaningful ambiguity from annotation noise. We argue that this perspective should guide how AI systems are represented, learned, evaluated, deployed, and governed, supporting more human-centered AI that better reflects the diversity of human interpretation.
\end{abstract}

\section{Introduction}

\begin{wrapfigure}{r}{0.34\textwidth} %0.43 - original
    \centering
    \vspace{-10pt}
    \includegraphics[width=\linewidth]{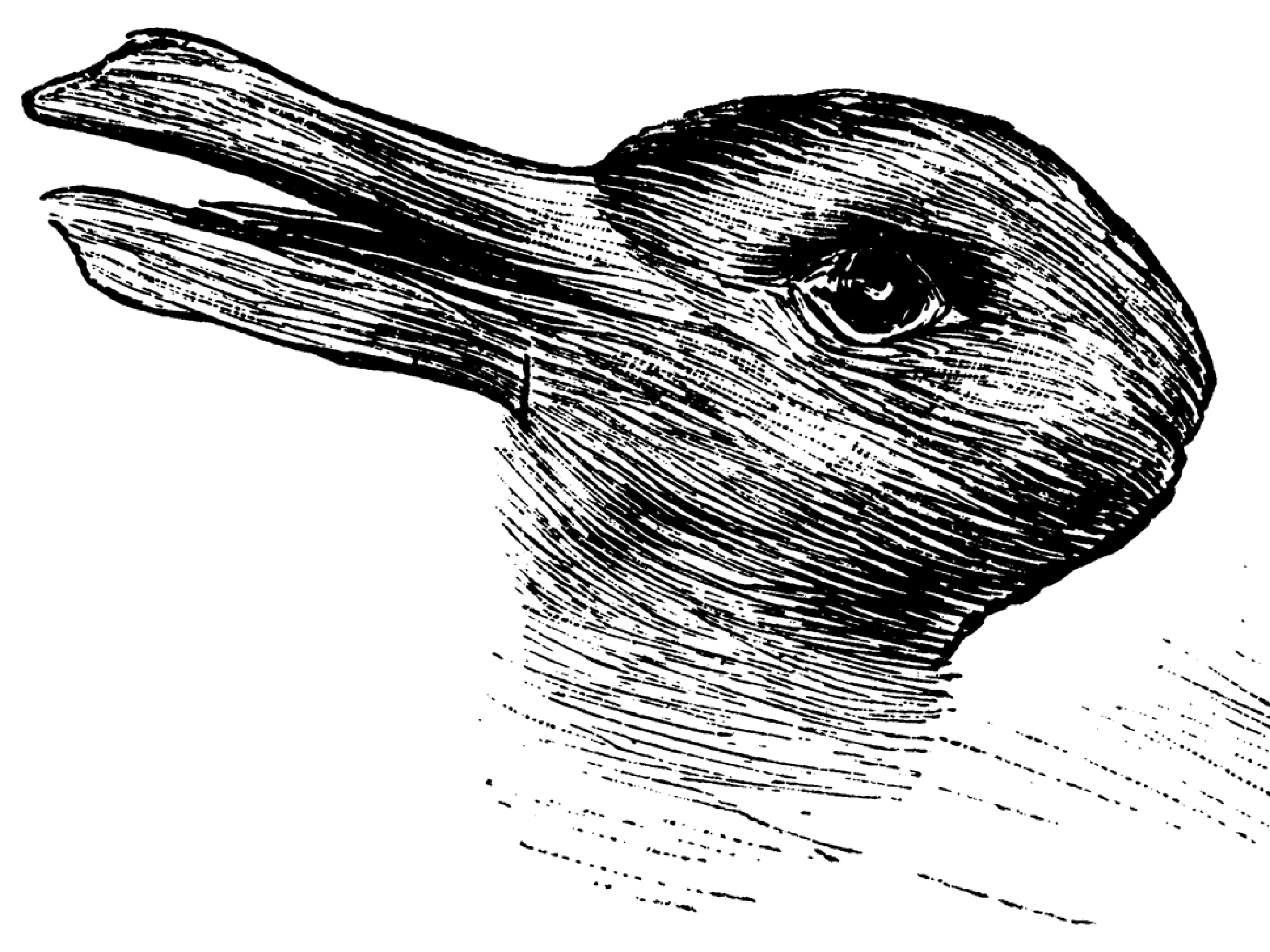}
    \caption{The Duck--Rabbit illusion illustrates ambiguity in human interpretation \cite{duckrabbit1892}.}
    \label{fig:duckrabbit}
    \vspace{-10pt}
\end{wrapfigure}

Current AI systems are advancing rapidly, with the next generation aiming to understand and interact with humans in increasingly natural ways. As artificial intelligence becomes increasingly embedded in everyday social environments, from healthcare and educational platforms to conversational agents and decision support systems, machines are no longer simply processing information but participating in human interaction.  In this emerging human-machine symbiotic society, AI systems must appropriately understand human expressions, behaviours, and interpretations to develop technologies that are trustworthy, socially compatible, and capable of meaningful interaction with people.
%In this emerging human machine symbiotic society, the ability of AI systems to appropriately understand human expressions, behaviours, and interpretations is increasingly important for developing technologies that are trustworthy, socially compatible, and capable of meaningful interaction with people.

However, understanding humans often involves \emph{ambiguity}. In many human-centered tasks, different individuals may interpret the same text, speech, or interaction differently depending on contextual understanding, personal experience, cultural background, or emotional perception~\cite{uma2021learning}. Figure~\ref{fig:duckrabbit} illustrates the classic Duck Rabbit illusion, where the same visual stimulus may be interpreted as either a duck or a rabbit. %Both interpretations are valid, even though the input remains unchanged. 
In this paper, \emph{ambiguity refers to variation among people’s interpretations of the same input, where multiple interpretations may be reasonable and valid.} We distinguish `ambiguity' from `model uncertainty' used in the broad machine learning literature, which reflects limitations in a model’s knowledge or confidence and may be reduced or calibrated, whereas ambiguity represents meaningful variation in human interpretation that should be preserved and learned~\cite{hullermeier2021aleatoric}. Such ambiguity is not necessarily caused by missing information or annotation error. Rather, it may arise from the inherent subjectivity of human perception and interpretation. %Human interpretations of emotion, behaviour, language, and social meaning are shaped by personal experience, cultural and educational background, job function, domain expertise, individual perspective, and situational context. 
For example, a smile may signal happiness, politeness, irony, or discomfort depending on the circumstances~\citep{wu2024emotion}. %\textcolor{blue}{Even the concept Artificial Intelligence have inherent ambiguity. One definition of Artificial Intelligence suggest that Artificial Intelligence is a process rather than a phenomenon referring to the frontier of computational advancements addressing decision making problems with human intelligence as point of reference~\cite{berente2021managing}.  In this definition the AI of yesterday, might not be considered as AI today, and the AI of today might not be considered as AI tomorrow. Other AI definitions has a regulatory purpose e.g. the EU AI Act~\cite{EU_DATA_ACT} where the definition has criteria. AI systems fulfilling the defined criteria is then within regulatory scope and must comply with specified requirements.}

% Berente, N., Gu, B., Recker, J., & Santhanam, R. (2021).Managing Artificial Intelligence1. MIS Quarterly 1 September 2021; 45 (3): 1433–1450. https://doi.org/10.25300/MISQ/2021/16274

% European Data Protection Supervisor, AI Act Regulation (EU) 2024/1689 – Regulation (EU) 2024/1689 of the European Parliament and of the Council of 13 June 2024 laying down harmonised rules on artificial intelligence and amending Regulations (EC) No 300/2008, (EU) No 167/2013, (EU) No 168/2013, (EU) 2018/858, (EU) 2018/1139 and (EU) 2019/2144 and Directives 2014/90/EU, (EU) 2016/797 and (EU) 2020/1828 (Artificial Intelligence Act) (Text with EEA relevance), Publications Office of the European Un-ion, 2025, https://data.europa.eu/doi/10.2804/4225375

This inherent ambiguity distinguishes many human centered tasks from traditional pattern recognition problems such as speech recognition or object identification, where ground truth is relatively well defined~\cite{baan2022stop}. For many human centered tasks, a single definitive label may not exist because several interpretations of the same input may all be reasonable~\cite{baan2022stop, nie2020can}. As a result, computational systems should not be expected to recover a precise label when such a label may not even exist. This raises a central challenge for artificial intelligence: \textit{how should AI systems understand and model subjective phenomena when no single, definitive ground truth exists?}

Generally, AI systems assume that each input should ultimately correspond to a single correct label. When multiple annotations are available, they are often aggregated through majority voting or averaging, with variation treated as noise or uncertainty to be reduced~\citep{sandri2023don}. However, in tasks involving human interpretation, such variation may arise from both \textbf{meaningful ambiguity}~\cite{wu2026shala, chen-etal-2025-rose} and \textbf{annotation noise}~\cite{klie2024analyzing}. While noise may result from errors, inattentiveness, or poor task design, ambiguity can reflect genuine differences in how the same input is interpreted across people, contexts, and domains. Collapsing these interpretations into a single target can therefore remove meaningful information.

Although recent work has begun to recognize ambiguity in AI and machine learning, existing approaches remain fragmented, often address only specific tasks or stages of the AI pipeline, and rarely distinguish meaningful ambiguity from annotation noise~\cite{ambinli, sethu2019ambiguous, wu2024emotion}. %\textbf{We therefore argue that human-centered AI should preserve and account for meaningful ambiguity rather than collapse diverse human interpretations into a single consensus.}
\textbf{We therefore argue that human-centered AI should embrace meaningful ambiguity rather than collapse diverse human interpretations into a single consensus, while explicitly distinguishing ambiguity from annotation noise.}

In this position paper, we present a unified perspective for considering ambiguity across the AI life cycle, including representation, learning, evaluation, deployment, maintenance, and governance. Through this perspective, we aim to establish shared principles for handling ambiguity in human-centered AI and support the development of systems that are more responsible, inclusive, and better suited to diverse human and societal contexts.

\section{Current AI Approaches to Ambiguity}

Annotations for human-centered tasks may vary across annotators, and this variation can contain important information rather than merely reflect annotation error~\cite{aroyo2015truth,davani2022dealing, chen-etal-2024-seeing, plank-2022-problem}. This phenomenon has been studied across multiple domains, including semantic reasoning tasks such as natural language inference~\cite{jayaweera2025disagreement, pavlick-kwiatkowski-2019-inherent}, and subjective perception tasks such as emotion recognition and affective computing~\cite{niu2025rethinking, chou-2025, wu2024dual, sethu2019ambiguous}. %Recent advances in AI have increasingly acknowledged the role of ambiguity in human-centered subjective tasks. 
A range of approaches have been proposed to move beyond single-label supervision, including multi-annotator modeling, distributional labels, uncertainty estimation, and annotator-aware learning. In categorical tasks, e.g., natural language inference and discrete emotion recognition, prior work has explored soft-label supervision~\cite{wu2026amber, fard2025affectnet}, multi-label formulations~\cite{ando19_interspeech, ju2020}, and ambiguity-aware learning approaches~\cite{chou2019every}. In continuous affect prediction, ambiguity has been represented using distributions (Gaussian~\cite{han2017hard}, Beta~\citep{wu2024dual,beta, wu2024can}, and non-parametric label distributions~\citep{wu2022novel}), mostly focusing on variance as a measure of uncertainty, where overlap measures ambiguity. More recently, language and multimodal models have increasingly explored ways to account for variability in human judgment, including through ambiguity-aware reinforcement learning~\cite{wu2026shala} and inference-time decoding strategies~\cite{jia2026decoding}. \textcolor{black}{Others have problematized treating ground truth labels as objective if the underlying knowledge is uncertain, they furthermore describe issues occurring from evaluating AI systems based on know-what measures when domain experts work is evaluated according to know-how~\cite{lebovitz2021ai}}. 
% Lebovitz, S., Levina, N. and Lifshitz-Assaf, H. (2021) ‘Is AI ground truth really “true”? The dangers of training and evaluating AI tools based on experts’ know-what’, Management Information Systems Quarterly.
Collectively, these developments represent important progress towards recognizing that human annotations are not always reducible to a single ground truth.

Despite this progress, current approaches remain largely method-driven and task-specific. Different representations, learning strategies, and evaluation protocols are often developed independently, resulting in inconsistent assumptions about how ambiguity should be modeled and utilized. More fundamentally, most existing methods focus on modeling the observed annotations rather than the underlying space of plausible human interpretations from which those annotations arise. Consequently, ambiguity is often treated as a property of individual datasets or algorithms, rather than as a fundamental characteristic of subjective AI tasks that should be considered consistently throughout AI development and deployment.

\section{Why Ambiguity Matters for Human-centered AI}\label{sec:limitation}
%\section{Related Work}\label{sec:limitation}

Ambiguity in human interpretation has implications beyond the annotations themselves. Whether variation reflects meaningful human interpretation or annotation noise determines what information an AI system should learn, while the interpretations it preserves can directly shape its subsequent decisions and actions. Understanding both aspects is therefore central to developing AI systems that appropriately operate under ambiguity.

\subsection{Ambiguity in Human Annotation}
%Annotations for human-centered tasks may vary across annotators, and this variation can contain important information rather than merely reflect annotation error \cite{aroyo2015truth,davani2022dealing, chen-etal-2024-seeing, plank-2022-problem}. This phenomenon has been studied across multiple domains, including semantic reasoning tasks such as natural language inference \cite{jayaweera2025disagreement, pavlick-kwiatkowski-2019-inherent}, and subjective perception tasks such as emotion recognition and affective computing \cite{niu2025rethinking, chou-2025, wu2024dual, sethu2019ambiguous}. 
Human perceptions and annotations naturally vary across different individuals, and importantly, not all variation among annotations has the same origin. Some variation reflects meaningful ambiguity, whereas other variation arises from annotation noise. Distinguishing between the two is essential for determining which signals AI systems should learn from and which should be mitigated.

\paragraph{Sources of Ambiguity.}
Ambiguity in human annotation can arise from different sources, including individual differences, contextual dependence, variation across domains and geographic regions, changes over time, and ambiguity in the input itself. Together, these factors can lead annotators to form different yet reasonable interpretations of the same input. Table~\ref{tab:sources_disagreement} summarizes the representative sources and examples.

\begin{table}[b!]
\centering
\caption{Sources of ambiguity and disagreement in human annotation.}
\label{tab:sources_disagreement}
\begin{tabularx}{\linewidth}{|p{0.18\linewidth}|p{0.32\linewidth}|X|}
\hline
\textbf{Category} & \textbf{Description} & \textbf{Example} \\
\hline

Individual differences~\citep{davani2022dealing,sap2022annotators,pei2023annotator} &
Annotators may interpret the same behavior differently because of differences in experience, culture, expertise, or prior knowledge. &
Avoiding eye contact may be interpreted as disrespectful by one annotator, but as a sign of respect, discomfort, or shyness by another. \\
\hline

Contextual dependence~\citep{pavlick2019inherent,sandri2023don} &
The interpretation of an input may depend on the immediate situation in which it occurs. &
A person speaking loudly may be interpreted as angry during a quiet meeting, but as enthusiastic at a sporting event. \\
\hline

Domain and geographic variation~\citep{davani2024d3code} &
The meaning of a concept or the appropriate judgment may differ across professional domains, organizations, countries, or regulatory jurisdictions. &
The same AI application may be classified as low risk in one domain or country but as high risk in another because different standards or regulations apply. \\
\hline

Temporal change~\citep{abercrombie2023temporal} &
Interpretations may change over time as individual context, knowledge, regulations, standards, or social norms evolve. &
%An AI system considered acceptable when it was introduced may require changes after new regulations or standards are adopted. \\
The same person may interpret the same message differently at different times as their context or knowledge changes.\\
\hline

Input ambiguity~\cite{aroyo2015truth,pavlick2019inherent} &
The input itself may support more than one reasonable interpretation. &
The same facial expression may indicate anger, stress, concentration, or discomfort. \\
\hline

\end{tabularx}
\end{table}

\paragraph{Sources of Noise.}
Annotation noise arises from imperfections in the annotation process rather than meaningful variation in human interpretation. It may result from annotation errors, low-quality or corrupted inputs, reduced attention or fatigue, and poorly designed instructions or \textcolor{black}{annotation interfaces~\citep{daniel2018quality,sandri2023don}.} Unlike ambiguity, which reflects multiple reasonable interpretations, noise represents unintended variation that should generally be reduced. Table~\ref{tab:sources_noise} summarizes the representative sources and examples of noise.

\paragraph{Ambiguity vs. Noise.}
Distinguishing ambiguity from annotation noise is essential. While noise should generally be minimized, ambiguity carries meaningful information about variation in human perception and interpretation. Treating all annotation variability as noise risks discarding valuable signals about how different individuals understand the same input. Recognizing ambiguity as an inherent characteristic of human annotation has important implications for AI systems. For subjective tasks, the goal should not be to eliminate variation in pursuit of a single ground truth, but to preserve and learn from meaningful ambiguity while \textcolor{black}{mitigating annotation noise~\citep{ivey2025nutmeg}.} This distinction motivates a closer examination of how current AI approaches represent, learn from, and evaluate ambiguity, which we discuss in Section~\ref{sec:system} .

% \paragraph{Ambiguity Vs. Noise.} The distinction between disagreement and noise is crucial. While noise should generally be minimized, disagreement carries meaningful information about the diversity of human perception and interpretation. Treating all variability as noise risks discarding valuable signals that reflect how different individuals understand the same input. Table 3 highlights the key differences between these two forms of variability. 

% Understanding disagreement as a structural property of human annotation has important implications for AI systems. It suggests that, for subjective tasks, the goal should not be to eliminate variability in pursuit of a single ground truth, but rather to account for and make use of this variability. This perspective motivates a closer examination of how current AI approaches handle disagreement, which we discuss in the next section. 

\begin{table}[tb]
\centering
\caption{Sources of noise in human annotation.}
\label{tab:sources_noise}
\begin{tabularx}{\linewidth}{|p{0.18\linewidth}|p{0.32\linewidth}|X|}
\hline
\textbf{Category} & \textbf{Description} & \textbf{Example} \\
\hline

Annotation error~\citep{raykar2010learning} &
Accidental mistakes made during the annotation process. &
An annotator selects ``angry'' instead of ``happy'' by mistake or clicks the wrong response option. \\
\hline

Low-quality input~\citep{hsueh2009data} &
The input is unclear, incomplete, or degraded, making reliable interpretation difficult. &
Background noise makes a speech recording difficult to understand, or poor lighting obscures a facial expression. \\
\hline

Attention and fatigue~\citep{tu2020attention} &
Reduced attention, tiredness, or loss of concentration leads to inconsistent annotations. &
After labeling many similar samples, an annotator begins responding quickly without carefully reviewing each one. \\
\hline

Task design issues~\citep{mathur2017sequence,oortwijn2021interrater} &
Unclear instructions, poorly defined labels, or interface limitations introduce avoidable inconsistencies. &
Annotators use the labels ``concerned'' and ``anxious'' differently because the distinction between them is not clearly explained. \\
\hline

\end{tabularx}
\end{table}

\subsection{Implications in Decision Making Under Ambiguity}

Such an AI system should include, where appropriate: the most likely interpretation, alternative plausible interpretations, the degree of ambiguity or uncertainty, the context in which the output is valid, and guidance on what kinds of decisions the output can and cannot support. 
This section explicitly discusses the ideas in different applications including affective computing, natural language understanding,  robotics, and medicine. 

\paragraph{Affective Computing.}  Consider a conversational AI system that detects a user's emotional state and adapts its response accordingly. Such systems learn from human perceptions of emotion, for which a single consensus interpretation may not always exist. For example, one person may perceive a user's expression as angry, while another may interpret the same expression as sarcastic.
An AI 
that sees only “angry” may respond defensively or escalate unnecessarily. 
On the contrary, an AI system that sees “could be angry, but also substantially sarcastic” may instead recognize that the person’s state is ambiguous and select a response that acknowledges both possibilities, such as de-escalating the tone while also asking clarifying questions.

\paragraph{Natural Language Understanding.} Consider a content moderation system that determines whether an online post is harmless, offensive, harassing, hateful, satirical, or politically critical.
These categories often depend on context, target group, local norms, and the identity or experience of the people interpreting the content. 
For example, most annotators may mark a post as non-offensive, while annotators from the targeted group may consistently perceive it as harmful. Such patterned disagreement is not annotation noise: 
it reflects a coherent perspective associated with how the content affects a particular group. In that case, the minority interpretation may be the most relevant signal for moderation.  

\paragraph{Robotics and planning.} Robotics provides another example where preserving multiple possible interpretations can directly affect safety~\citep{hoffman2024inferring}. A robot working near a human may need to infer whether the person intends to hand over an object, move away, ask for help, or stop the interaction. 
For example, a person reaching toward an object may be intending to hand it to the robot, reposition it, or take it back. 
The decision value of the output is not only the most likely intention, but whether the ambiguity is low enough to justify physical action.
Such ambiguity should be distinguished from transient sensor errors or corrupted observations, which should be filtered out.  

%\paragraph{Medicine.}
%In clinical trials, symptoms and adverse events must be assessed to determine whether they are related to the treatment or to other factors. For example, elevated blood pressure may result from the investigational treatment, an underlying condition, another medication, or normal physiological variation. The measurement itself may be clear, but its causal attribution can remain ambiguous because several explanations are plausible.

\paragraph{Healthcare.} In healthcare, the same patient information may be interpreted differently by different clinicians. For example, clinicians reviewing the same patient-reported symptoms may reasonably differ in their assessment of symptom severity or clinical significance based on their expertise, experience, and contextual understanding. Collapsing these assessments into a single label may obscure meaningful differences in clinical interpretation. Ambiguity-aware AI could instead preserve such variation and communicate multiple plausible interpretations, supporting more transparent and responsible clinical decision-making.

Across these examples, the central point is that AI outputs should be designed for the decisions they are meant to support. In some cases, a single label or point estimate may be sufficient. In others, it may hide information that is important for human judgment, such as disagreement among annotators, multiple plausible risk outcomes, uncertain human intention, or context-dependent interpretations of harmful content. 
Decisions should be made with an understanding of what the model output represents and what it leaves unresolved.

\section{A Conceptual Framework for AI Systems to Learn Under Ambiguity}\label{sec:system}

Building on the observations in Section~\ref{sec:limitation}, we argue that subjective tasks should be understood through an interpretation space rather than a single ground-truth label. Instead of treating ambiguity as noise to be removed, we view it as an informative signal that reflects the diversity of human judgments, perceptions, and experiences. This perspective shifts the objective of AI systems from predicting a single correct answer to modeling the space of plausible human interpretations while distinguishing meaningful ambiguity from annotation noise. 

Let $A(x)={y_1,y_2,\ldots,y_K}$ denote the set of annotations collected for an input stimulus (x). We conceptually distinguish the observed annotations into two components: 
\begin{equation}
    A(x)=S(x)\oplus N(x) ,
\end{equation}
where $S(x)$ denotes the interpretation space containing meaningful human ambiguity and disagreement. $N(x)$ denotes annotation noise arising from factors such as annotation errors, low-quality stimuli, or inattentive labeling, and $\oplus$ conceptually denotes that the observed annotations contain both components. We emphasize that this decomposition is conceptual rather than algorithmic; the objective is not to prescribe how $S(x)$ and $N(x)$ should be separated, but to recognize that they should be treated differently throughout the AI pipeline. 

\begin{figure*}[!tb]
\centering
\includegraphics[width=\linewidth, height = 6.7cm]{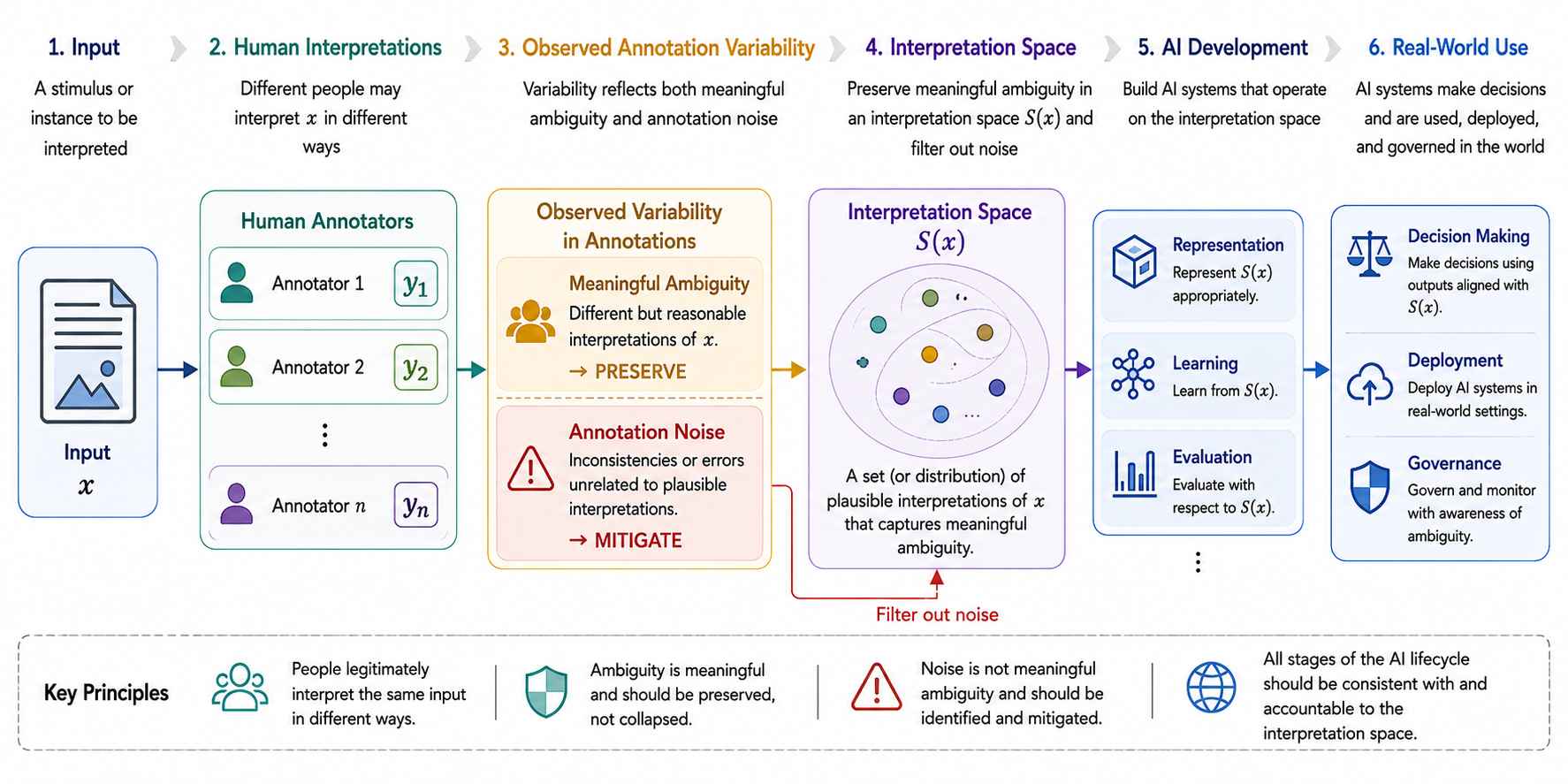}
\caption{Overview of the proposed conceptual framework for developing human-centered AI under ambiguity.}
\label{fig:framework}
\end{figure*}

Under this framework, the development of AI systems for subjective tasks can be viewed through three interconnected components: representation, learning, and evaluation. The following sections discuss how ambiguity and annotation noise should be considered across each stage of the AI pipeline. Fig.~\ref{fig:framework} depicts an overview of the proposed conceptual framework. 

%\vspace{-1em}

\subsection{Representation: Preserving the Ambiguity in Interpretation Space}

\emph{We envision representation paradigms that preserve meaningful ambiguity while explicitly distinguishing them from annotation noise, enabling AI systems to operate on the full interpretation space of human judgments rather than a single consensus outcome.} 

Conceptually, the representation process can be expressed as: 
\begin{equation}
    R(A(x)) = R\big(S(x), N(x)\big),
\end{equation} 
where $R(\cdot)$ denotes a representation of the observed annotations, preserving information regarding both the interpretation space $S(x)$ and annotation noise $N(x)$. 

\paragraph{Preserve interpretive diversity.} The interpretation space may be represented in different forms, including distributions of annotations, sets of valid labels, preference structures, annotator-specific representations, or latent representations that capture interpretive variability. While no single representation is universally appropriate, the key principle is that representations should retain information regarding the diversity of human judgments rather than reduce them to a single consensus outcome. 

\paragraph{Distinguish ambiguity from annotation noise.} A fundamental challenge of human-centered AI is not only preserving inherent label ambiguity, but also distinguishing meaningful ambiguity from annotation noise. Annotation variability arising from differences in perception, context, experience, or cultural background should be retained as part of the interpretation space, whereas variability introduced by annotation errors, inattentive labeling, or poor stimulus quality should be explicitly identified and treated separately. Although we do not prescribe a specific mechanism for this distinction, future representation paradigms should explicitly account for both components, providing a principled foundation for downstream learning, evaluation, and governance.

\subsection{Model Training: Learning from the Interpretation Space}

\emph{We call for a redefinition of what it means to "learn" in subjective AI: not recovering a single consensus label, but learning from the interpretation space while remaining robust to annotation noise.}

Conceptually, ambiguity-aware learning can be formulated as: 
\begin{equation}
\theta^*=\arg\min_{\theta}\mathcal{L}\big(f(x,\theta),S(x),N(x)\big) ,
\end{equation}
where optimization is guided by the interpretation space (S(x)) while accounting for annotation noise (N(x)).

\paragraph{Learn from meaningful ambiguity.} Learning objectives should treat ambiguity and disagreement as informative supervision signals rather than annotation artifacts. Instead of encouraging models to converge toward a single dominant interpretation, optimization should leverage the diversity contained within the interpretation space to better capture the complexity and subjectivity of human-centered tasks. 

\paragraph{Remain robust to annotation noise.} While ambiguity reflects meaningful differences in human interpretation, annotation noise does not provide useful supervision. Future learning paradigms should therefore be designed to distinguish informative ambiguity from annotation artifacts, allowing models to learn from genuine interpretive diversity while reducing the influence of noisy annotations. We do not prescribe a specific optimization strategy, but argue that both components should be explicitly considered throughout the learning process.

\subsection{Model Evaluation: Beyond "Ground Truth" Accuracy}  

\emph{We advocate for new evaluation protocols that reward models not for converging to a single consensus answer, but for faithfully capturing the interpretation space while remaining robust to annotation noise.}

Conceptually, evaluation can be formulated as: 
\begin{equation}
  E\big(f_{\theta}(x),S(x),N(x)\big) ,
\end{equation}
where $E(\cdot)$ measures the degree to which model outputs align with the interpretation space while accounting for annotation noise. 

\paragraph{Evaluate beyond correctness.} No single metric is universally appropriate for subjective tasks. Depending on how the interpretation space is represented, evaluation may consider alignment with distributions of judgments, consistency with multiple valid labels, preservation of interpretive variability, or agreement with preference structures. The objective of evaluation should therefore extend beyond deterministic correctness and instead assess how faithfully a model captures the diversity of human interpretations. 

\paragraph{Evaluate consistency with subjectivity.} Evaluation should remain aligned with the assumptions adopted during representation and learning. Models designed to preserve and learn from the interpretation space should be evaluated with respect to the same interpretation space, while minimizing the influence of annotation noise. Such consistency ensures that ambiguity is treated as meaningful information throughout the entire AI pipeline rather than being inadvertently discarded during evaluation.

\section{Governing \textcolor{black}{Human-Centered} AI Systems }

\emph{We call for action: AI systems should not simply determine what is true. Instead, they should account for what interpretations are plausible, how strongly they are supported, where ambiguity exists, and what actions are appropriate under that ambiguity.}

%The framework proposed in this paper has implications extending beyond model development. If subjective AI systems are designed to represent an interpretation space rather than represent a single deterministic ground truth, then their governance must also evolve.
Conventional AI governance largely assumes that an AI system approximates an objective truth and that system quality can therefore be assessed by correctness, accuracy, or agreement with a reference label or a performance benchmark. While these assumptions are appropriate for many prediction tasks, they become insufficient for inherently subjective problems where multiple interpretations may simultaneously be legitimate. 

Consequently, the objective of governance is no longer to ensure that an AI system consistently produces the correct answer. Rather, \emph{governance should ensure that the system faithfully represents legitimate interpretive diversity, distinguishes meaningful disagreement from annotation noise, communicates ambiguity appropriately, and supports human decision-making within established ethical and regulatory boundaries.} %\textcolor{black}{Importantly, preserving multiple plausible interpretations does not imply that all interpretations or resulting actions are equally acceptable. In other words, ambiguity does imply ethical relativism; how AI systems act upon ambiguity remains constrained by ethical principles, legal requirements, and domain-specific standards.} %\jw{I added this sentence which is meant to include the ideas section 5.4 was trying to convey, and then I think we can remove section 5.4.}

This perspective fundamentally changes how human-centered AI systems should be validated, deployed, monitored, and maintained throughout their life cycle. 

\subsection{Validation Beyond Deterministic Ground Truth} 

For objective tasks, validation evaluates agreement between model predictions and a reference truth (benchmarking). For subjective tasks, validation should instead evaluate whether the model appropriately represents the interpretation space associated with the intended deployment context. 

Rather than asking whether a prediction is correct, governance should ask whether the model preserves legitimate diversity of interpretation, distinguishes disagreement from annotation noise, remains calibrated under ambiguity, behaves consistently across relevant populations and contexts of use, and communicates uncertainty in ways that support appropriate human decisions. 

Importantly, acceptable ambiguity depends on application risk. Low-risk recommendation systems may tolerate substantial interpretive variability, whereas high-risk applications, including healthcare, pharmaceutical development, finance, and legal decision support, require considerably narrower operational boundaries, stronger evidence, and increased human oversight. Governance should therefore adopt a risk-based perspective in which acceptable ambiguity is determined by the potential consequences of incorrect interpretation rather than by technical model performance alone. 

\subsection{Human Oversight Under Ambiguity}

Traditional human-in-the-loop systems frequently assume that humans resolve residual uncertainty by selecting the correct answer or by course-correcting the AI system in question. In human-centered AI, however, disagreement may itself represent the appropriate outcome. Human oversight therefore shifts from identifying a single correct prediction to interpreting multiple plausible explanations within their operational context. Indeed, ambiguity may even arise between the outcome of the AI system and the human-in-the-loop, in a way where it becomes unclear to the human if course-correction is needed or not, due to the subjectivity of the task at hand. 

Consequently, human-centered AI systems should avoid presenting outputs as definitive whenever multiple interpretations remain plausible. Instead, systems should communicate the most likely interpretation,  the confidence in the given interpretation, alternative plausible interpretations, the degree of disagreement, contextual assumptions underlying each interpretation; and situations in which additional evidence or human review is required.  

Such communication enables decision-makers to reason about ambiguity explicitly rather than implicitly assuming certainty. This may reduce inappropriate automation, improve transparency, and support more calibrated trust in AI-assisted decision-making.

\subsection{Life Cycle Monitoring and Interpretation Spaces}
Deployment does not eliminate subjectivity. Human interpretation evolves over time as cultures, organizations, regulations, and professional practices change. Traditional machine learning operations therefore requires extension beyond concept drift and data drift to include subjectivity drift: changes in legitimate human interpretations despite relatively stable input data.

Governance should monitor interpretation spaces throughout the system life cycle through mechanisms such as re-annotation, stakeholder feedback, expert review, and calibration exercises. Importantly, governance must distinguish evolving ambiguity from annotation noise, as these require fundamentally different responses. Changes in human interpretation may justify updating models, whereas noise primarily motivates improvements in data quality, annotation procedures, and quality assurance. Accordingly, governance should evaluate whether AI systems maintain an appropriate representation of the interpretation space over time rather than whether they collapse ambiguity into consensus.

\subsection{ \textcolor{black}{Ambiguity} Does Not Imply Ethical Relativism }
%\jw{could we remove this section for the sake of saving space?}
%It is important to note that subjective AI does not lead to ethical relativism. 
It is important to clarify that accounting for subjectivity and ambiguity in AI does not imply ethical relativism. Throughout this paper we argue that many \textcolor{black}{human-centered} AI tasks lack a single \textcolor{black}{definitive ground truth} because human interpretation \textcolor{black}{inherently} varies across individuals owing to different contexts, expertise, cultures and other factors. This observation concerns how AI systems should model empirical human judgments. It does not imply that ethical principles themselves are subjective or that all interpretations should be treated as equally acceptable. 

Instead, we distinguish three fundamentally different concepts: 

\begin{itemize}[nosep]
    \item \textbf{Interpretation space}, representing legitimate diversity in human judgment; 
    \item \textbf{Annotation noise}, representing unintended errors and artifacts that should be reduced; and  
    \item \textbf{Normative constraints}, representing ethical principles, legal obligations, organizational policies, and domain-specific standards that govern how AI systems ought to behave.  
\end{itemize}

Ethics therefore exists outside the interpretation space. \textcolor{black}{Rater} {ambiguity} describes how people interpret the world while ethical governance determines how AI systems should act upon those interpretations. 

Indeed, preserving \textcolor{black}{ambiguity} may strengthen ethical AI rather than weaken it. Minority perspectives frequently reveal systematic harms, cultural blind spots, demographic disparities, or emerging risks that majority-vote aggregation may obscure. Maintaining these perspectives enables organizations to identify potentially vulnerable populations, recognize fairness concerns earlier, and support more inclusive decision-making while remaining fully constrained by established ethical principles and regulatory requirements.

\subsection{Design Principles for Governing Human-centered AI}

%\jw{I think the principles are great! My concern is, some of the content are repeated from previous subsections in sec5, is there a better way to remove the repetitions since we are out of space? } Comment from Per: {We could consider using gregor et al. 2020 (https://web.archive.org/web/20210121042121id_/https://aisel.aisnet.org/cgi/viewcontent.cgi?article=1978&context=jais) to strucutre the articulation of our design principles}

The preceding discussion suggests seven principles for governing AI systems operating under inherent human ambiguity. 

\paragraph{Principle 1.} Validate interpretation spaces rather than deterministic labels. Evaluation should assess whether model behavior faithfully represents legitimate human interpretation rather than maximizing agreement with a single aggregated label. 

\paragraph{Principle 2.} Distinguish disagreement from annotation noise. Governance should preserve meaningful interpretive diversity while continuously reducing errors arising from poor-quality data or annotation artifacts. 

\paragraph{Principle 3.} Govern ambiguity according to application risk. Acceptable ambiguity should depend on the potential consequences of downstream decisions, with progressively stronger controls required as potential harm increases. 

\paragraph{Principle 4.} Monitor subjectivity throughout the system life cycle. Governance should monitor evolving human interpretations alongside conventional indicators such as concept drift and model degradation. 

\paragraph{Principle 5.} Communicate ambiguity rather than conceal it. AI systems should transparently communicate multiple plausible interpretations and their associated levels of disagreement whenever this information is relevant for decision-making. 

\paragraph{Principle 6.} Separate empirical subjectivity from normative governance. AI systems should faithfully represent legitimate human ambiguity while remaining constrained by ethical principles, regulatory requirements, and organizational values. 

\paragraph{Principle 7.} Do not consider human-centered AI systems as isolated entities during monitoring, evaluation, and retraining but as sociotechnical systems interacting with users, IT infrastructure, and other AI systems or AI agents through both input and output potentially with multiple purposes.

These principles extend AI governance beyond deterministic notions of correctness toward governance mechanisms capable of supporting AI systems operating under inherent human subjectivity. Rather than seeking to eliminate disagreement, subjective AI should enable organizations to reason about disagreement responsibly, transparently, and ethically throughout the entire AI life cycle.

\section{Conclusion}

%Ambiguity is an inherent part of human interpretation and should be treated as a meaningful consideration in the development of human-centered AI. In this paper, we explicitly distinguish meaningful ambiguity in human annotation from annotation noise, and propose a general framework for representing, learning from, and evaluating the space of plausible human interpretations. We further call for ambiguity to be considered throughout AI decision making, deployment, and governance. Through this position, we hope to encourage a broader shift towards AI systems that better reflect the diversity and complexity of human interpretation and interact with people in more appropriate and responsible ways.

In this position paper, we argue that ambiguity is an inherent part of human interpretation and should be explicitly considered in the development of human-centered AI. We propose a general conceptual framework that distinguishes meaningful ambiguity from annotation noise and models multiple plausible human interpretations through an interpretation space. Building on this framework, we call for action to account for ambiguity throughout AI representation, learning, evaluation, decision making, deployment, and governance. We hope this perspective provides a foundation for rethinking how AI systems learn from human interpretations and paves the way towards more human-centered and responsible AI.

\section*{Acknowledgment}

Jingyao Wu and Mohammad Tariqul Islam are supported by MIT-Novo Nordisk Artificial Intelligence Postdoctoral Fellowship.

\bibliographystyle{unsrtnat}
\bibliography{mybib}

@misc{duckrabbit1892,
  author       = {{Anonymous}},
  title        = {Welche Thiere gleichen einander am meisten? Kaninchen und Ente},
  howpublished = {Fliegende Blätter, No. 2465},
  year         = {1892},
  note         = {Published 23 October 1892}
}

@inproceedings{wu2024emotion,
  title={Emotion recognition systems must embrace ambiguity},
  author={Wu, Jingyao and Dang, Ting and Sethu, Vidhyasaharan and Ambikairajah, Eliathamby},
  booktitle={2024 12th International Conference on Affective Computing and Intelligent Interaction Workshops and Demos (ACIIW)},
  pages={166--170},
  year={2024},
  organization={IEEE}
}

@article{davani2022dealing,
  title={Dealing with disagreements: Looking beyond the majority vote in subjective annotations},
  author={Davani, Aida Mostafazadeh and D{\'\i}az, Mark and Prabhakaran, Vinodkumar},
  journal={Transactions of the Association for Computational Linguistics},
  volume={10},
  pages={92--110},
  year={2022},
  publisher={MIT Press One Rogers Street, Cambridge, MA 02142-1209, USA journals-info~…}
}

@article{hoffman2024inferring,
  title={Inferring human intent and predicting human action in human--robot collaboration},
  author={Hoffman, Guy and Bhattacharjee, Tapomayukh and Nikolaidis, Stefanos},
  journal={Annual Review of Control, Robotics, and Autonomous Systems},
  volume={7},
  number={1},
  pages={73--95},
  year={2024},
  publisher={Annual Reviews}
}

@inproceedings{davani2024d3code,
  title={D3CODE: Disentangling disagreements in data across cultures on offensiveness detection and evaluation},
  author={Davani, Aida and Diaz, Mark and Baker, Dylan and Prabhakaran, Vinodkumar},
  booktitle={Proceedings of the 2024 Conference on Empirical Methods in Natural Language Processing},
  pages={18511--18526},
  year={2024}
}

@article{pavlick2019inherent,
  title={Inherent disagreements in human textual inferences},
  author={Pavlick, Ellie and Kwiatkowski, Tom},
  journal={Transactions of the Association for Computational Linguistics},
  volume={7},
  pages={677--694},
  year={2019},
  publisher={MIT Press One Rogers Street, Cambridge, MA 02142-1209, USA journals-info~…}
}

@inproceedings{sap2022annotators,
  title={Annotators with attitudes: How annotator beliefs and identities bias toxic language detection},
  author={Sap, Maarten and Swayamdipta, Swabha and Vianna, Laura and Zhou, Xuhui and Choi, Yejin and Smith, Noah A},
  booktitle={Proceedings of the 2022 conference of the north american chapter of the association for computational linguistics: Human language technologies},
  pages={5884--5906},
  year={2022}
}

@inproceedings{pei2023annotator,
  title={When do annotator demographics matter? measuring the influence of annotator demographics with the POPQUORN dataset},
  author={Pei, Jiaxin and Jurgens, David},
  booktitle={Proceedings of the 17th linguistic annotation workshop (LAW-XVII)},
  pages={252--265},
  year={2023}
}

@inproceedings{sandri2023don,
  title={Why don’t you do it right? analysing annotators’ disagreement in subjective tasks},
  author={Sandri, Marta and Leonardelli, Elisa and Tonelli, Sara and Je{\v{z}}ek, Elisabetta},
  booktitle={Proceedings of the 17th Conference of the European Chapter of the Association for Computational Linguistics},
  pages={2428--2441},
  year={2023}
}

@inproceedings{abercrombie2023temporal,
  title={Temporal and second language influence on intra-annotator agreement and stability in hate speech labelling},
  author={Abercrombie, Gavin and Hovy, Dirk and Prabhakaran, Vinodkumar},
  booktitle={Proceedings of the 17th linguistic annotation workshop (LAW-XVII)},
  pages={96--103},
  year={2023}
}

@article{uma2021learning,
  title={Learning from disagreement: A survey},
  author={Uma, Alexandra N and Fornaciari, Tommaso and Hovy, Dirk and Paun, Silviu and Plank, Barbara and Poesio, Massimo},
  journal={Journal of Artificial Intelligence Research},
  volume={72},
  pages={1385--1470},
  year={2021}
}

@article{raykar2010learning,
  title={Learning from crowds.},
  author={Raykar, Vikas C and Yu, Shipeng and Zhao, Linda H and Valadez, Gerardo Hermosillo and Florin, Charles and Bogoni, Luca and Moy, Linda},
  journal={Journal of machine learning research},
  volume={11},
  number={4},
  year={2010}
}

@inproceedings{hsueh2009data,
  title={Data quality from crowdsourcing: a study of annotation selection criteria},
  author={Hsueh, Pei-Yun and Melville, Prem and Sindhwani, Vikas},
  booktitle={Proceedings of the NAACL HLT 2009 workshop on active learning for natural language processing},
  pages={27--35},
  year={2009}
}

@inproceedings{tu2020attention,
  title={Attention-aware answers of the crowd},
  author={Tu, Jingzheng and Yu, Guoxian and Wang, Jun and Domeniconi, Carlotta and Zhang, Xiangliang},
  booktitle={Proceedings of the 2020 SIAM International Conference on Data Mining},
  pages={451--459},
  year={2020},
  organization={SIAM}
}

@inproceedings{mathur2017sequence,
  title={Sequence effects in crowdsourced annotations},
  author={Mathur, Nitika and Baldwin, Timothy and Cohn, Trevor},
  booktitle={Proceedings of the 2017 conference on empirical methods in natural language processing},
  pages={2860--2865},
  year={2017}
}

@inproceedings{oortwijn2021interrater,
  title={Interrater disagreement resolution: A systematic procedure to reach consensus in annotation tasks},
  author={Oortwijn, Yvette and Ossenkoppele, Thijs and Betti, Arianna},
  booktitle={Proceedings of the workshop on human evaluation of NLP systems (HumEval)},
  pages={131--141},
  year={2021}
}

@article{daniel2018quality,
  title={Quality control in crowdsourcing: A survey of quality attributes, assessment techniques, and assurance actions},
  author={Daniel, Florian and Kucherbaev, Pavel and Cappiello, Cinzia and Benatallah, Boualem and Allahbakhsh, Mohammad},
  journal={ACM Computing Surveys (CSUR)},
  volume={51},
  number={1},
  pages={1--40},
  year={2018},
  publisher={ACM New York, NY, USA}
}

@inproceedings{ivey2025nutmeg,
  title={Nutmeg: Separating signal from noise in annotator disagreement},
  author={Ivey, Jonathan and Gauch, Susan and Jurgens, David},
  booktitle={Proceedings of the 2025 Conference on Empirical Methods in Natural Language Processing},
  pages={2874--2887},
  year={2025}
}

@article{lebovitz2021ai,
  title={Is AI ground truth really true? The dangers of training and evaluating AI tools based on experts’ know-what},
  author={Lebovitz, Sarah and Levina, Natalia and Lifshitz-Assaf, Hila},
  journal={MIS quarterly},
  volume={45},
  number={3},
  pages={1501--1526},
  year={2021},
  publisher={Management Information Systems Research Center, University of Minnesota}
}

@inproceedings{chen-etal-2025-rose,
    title = "A Rose by Any Other Name: {LLM}-Generated Explanations Are Good Proxies for Human Explanations to Collect Label Distributions on {NLI}",
    author = "Chen, Beiduo  and
      Peng, Siyao  and
      Korhonen, Anna  and
      Plank, Barbara",
    editor = "Che, Wanxiang  and
      Nabende, Joyce  and
      Shutova, Ekaterina  and
      Pilehvar, Mohammad Taher",
    booktitle = "Findings of the Association for Computational Linguistics: ACL 2025",
    month = jul,
    year = "2025",
    address = "Vienna, Austria",
    publisher = "Association for Computational Linguistics",
    url = "https://aclanthology.org/2025.findings-acl.562/",
    doi = "10.18653/v1/2025.findings-acl.562",
    pages = "10777--10802",
    ISBN = "979-8-89176-256-5"
}

@inproceedings{chen-etal-2024-seeing,
    title = "``Seeing the Big through the Small'': Can {LLM}s Approximate Human Judgment Distributions on {NLI} from a Few Explanations?",
    author = "Chen, Beiduo  and
      Wang, Xinpeng  and
      Peng, Siyao  and
      Litschko, Robert  and
      Korhonen, Anna  and
      Plank, Barbara",
    editor = "Al-Onaizan, Yaser  and
      Bansal, Mohit  and
      Chen, Yun-Nung",
    booktitle = "Findings of the Association for Computational Linguistics: EMNLP 2024",
    month = nov,
    year = "2024",
    address = "Miami, Florida, USA",
    publisher = "Association for Computational Linguistics",
    url = "https://aclanthology.org/2024.findings-emnlp.842/",
    doi = "10.18653/v1/2024.findings-emnlp.842",
    pages = "14396--14419"
}

@inproceedings{jayaweera2025disagreement,
  title={From disagreement to understanding: The case for ambiguity detection in NLI},
  author={Jayaweera, Chathuri and Dorr, Bonnie},
  booktitle={Proceedings of the The 4th Workshop on Perspectivist Approaches to NLP},
  pages={37--46},
  year={2025}
}

@inproceedings{baan2022stop,
  title={Stop measuring calibration when humans disagree},
  author={Baan, Joris and Aziz, Wilker and Plank, Barbara and Fernandez, Raquel},
  booktitle={Proceedings of the 2022 Conference on Empirical Methods in Natural Language Processing},
  pages={1892--1915},
  year={2022}
}

@article{aroyo2015truth,
  title={Truth is a lie: Crowd truth and the seven myths of human annotation},
  author={Aroyo, Lora and Welty, Chris},
  journal={AI magazine},
  volume={36},
  number={1},
  pages={15--24},
  year={2015}
}

@article{jia2026decoding,
  title={Decoding Ambiguous Emotions with Test-Time Scaling in Audio-Language Models},
  author={Jia, Hong and Li, Weibin and Wu, Jingyao and Yu, Xiaofeng and Gao, Yan and Cheng, Jintao and Tang, Xiaoyu and Xia, Feng and Dang, Ting},
  journal={arXiv preprint arXiv:2602.03873},
  year={2026}
}

@article{wu2026amber,
  title={AmbER2: Dual Ambiguity-Aware Emotion Recognition Applied to Speech and Text},
  author={Wu, Jingyao and Lin, Grace and Song, Yinuo and Picard, Rosalind},
  journal={arXiv preprint arXiv:2601.18010},
  year={2026}
}

@article{pavlick-kwiatkowski-2019-inherent,
    title = "Inherent Disagreements in Human Textual Inferences",
    author = "Pavlick, Ellie  and
      Kwiatkowski, Tom",
    editor = "Lee, Lillian  and
      Johnson, Mark  and
      Roark, Brian  and
      Nenkova, Ani",
    journal = "Transactions of the Association for Computational Linguistics",
    volume = "7",
    year = "2019",
    address = "Cambridge, MA",
    publisher = "MIT Press",
    url = "https://aclanthology.org/Q19-1043/",
    doi = "10.1162/tacl_a_00293",
    pages = "677--694"
}

@article{chou-2025,
  author={Chou, Huang-Cheng and Goncalves, Lucas and Leem, Seong-Gyun and Salman, Ali N. and Lee, Chi-Chun and Busso, Carlos},
  journal={IEEE Transactions on Affective Computing}, 
  title={Minority Views Matter: Evaluating Speech Emotion Classifiers With Human Subjective Annotations by an All-Inclusive Aggregation Rule}, 
  year={2025},
  volume={16},
  number={1},
  pages={41-55},
  doi={10.1109/TAFFC.2024.3411290}}

@inproceedings{plank-2022-problem,
    title = "The ``Problem'' of Human Label Variation: On Ground Truth in Data, Modeling and Evaluation",
    author = "Plank, Barbara",
    editor = "Goldberg, Yoav  and
      Kozareva, Zornitsa  and
      Zhang, Yue",
    booktitle = "Proceedings of the 2022 Conference on Empirical Methods in Natural Language Processing",
    month = dec,
    year = "2022",
    address = "Abu Dhabi, United Arab Emirates",
    publisher = "Association for Computational Linguistics",
    url = "https://aclanthology.org/2022.emnlp-main.731/",
    doi = "10.18653/v1/2022.emnlp-main.731",
    pages = "10671--10682"
}

@inproceedings{ambinli,
    title = "Embracing Ambiguity: {S}hifting the Training Target of {NLI} Models",
    author = "Meissner, Johannes Mario  and
      Thumwanit, Napat  and
      Sugawara, Saku  and
      Aizawa, Akiko",
    booktitle = "Proceedings of the 59th Annual Meeting of the Association for Computational Linguistics and the 11th International Joint Conference on Natural Language Processing (Volume 2: Short Papers)",
    month = aug,
    year = "2021",
    address = "Online",
    publisher = "Association for Computational Linguistics",
    url = "https://aclanthology.org/2021.acl-short.109",
    doi = "10.18653/v1/2021.acl-short.109",
    pages = "862--869",
}

@article{beta,
author = {Bose, Deboshree and Sethu, Vidhyasaharan and Ambikairajah, Eliathamby},
title = {Continuous Emotion Ambiguity Prediction: Modeling With Beta Distributions},
year = {2024},
issue_date = {July-Sept. 2024},
publisher = {IEEE Computer Society Press},
address = {Washington, DC, USA},
volume = {15},
number = {3},
issn = {1949-3045},
url = {https://doi.org/10.1109/TAFFC.2024.3367371},
doi = {10.1109/TAFFC.2024.3367371},
journal = {IEEE Trans. Affect. Comput.},
month = jul,
pages = {1684–1695},
numpages = {12}
}

@inproceedings{ando19_interspeech,
  title     = {{Speech Emotion Recognition Based on Multi-Label Emotion Existence Model}},
  author    = {Atsushi Ando and Ryo Masumura and Hosana Kamiyama and Satoshi Kobashikawa and Yushi Aono},
  year      = {2019},
  booktitle = {{Interspeech 2019}},
  pages     = {2818--2822},
  doi       = {10.21437/Interspeech.2019-2524},
  issn      = {2958-1796},
}

@inproceedings{ju2020,
author = {Ju, Xincheng and Zhang, Dong and Li, Junhui and Zhou, Guodong},
title = {Transformer-based Label Set Generation for Multi-modal Multi-label Emotion Detection},
year = {2020},
isbn = {9781450379885},
publisher = {Association for Computing Machinery},
address = {New York, NY, USA},
url = {https://doi.org/10.1145/3394171.3413577},
doi = {10.1145/3394171.3413577},
booktitle = {Proceedings of the 28th ACM International Conference on Multimedia},
pages = {512–520},
numpages = {9},
location = {Seattle, WA, USA},
series = {MM '20}
}

@article{sethu2019ambiguous,
  title={The ambiguous world of emotion representation},
  author={Sethu, Vidhyasaharan and Provost, Emily Mower and Epps, Julien and Busso, Carlos and Cummins, Nicholas and Narayanan, Shrikanth},
  journal={arXiv preprint arXiv:1909.00360},
  year={2019}
}

@article{wu2024dual,
  title={Dual-constrained dynamical neural odes for ambiguity-aware continuous emotion prediction},
  author={Wu, Jingyao and Dang, Ting and Sethu, Vidhyasaharan and Ambikairajah, Eliathamby},
  journal={arXiv preprint arXiv:2407.21344},
  year={2024}
}

@inproceedings{chou2019every,
  title={Every rating matters: Joint learning of subjective labels and individual annotators for speech emotion classification},
  author={Chou, Huang-Cheng and Lee, Chi-Chun},
  booktitle={ICASSP 2019-2019 IEEE International Conference on Acoustics, Speech and Signal Processing (ICASSP)},
  pages={5886--5890},
  year={2019},
  organization={IEEE}
}

@article{niu2025rethinking,
  title={Rethinking emotion annotations in the era of large language models},
  author={Niu, Minxue and El-Tawil, Yara and Romana, Amrit and Provost, Emily Mower},
  journal={IEEE transactions on affective computing},
  year={2025},
  publisher={IEEE}
}

@article{fard2025affectnet,
  title={Affectnet+: A database for enhancing facial expression recognition with soft-labels},
  author={Fard, Ali Pourramezan and Hosseini, Mohammad Mehdi and Sweeny, Timothy D and Mahoor, Mohammad H},
  journal={IEEE Transactions on Affective Computing},
  year={2025},
  publisher={IEEE}
}

@inproceedings{han2017hard,
  title={From hard to soft: Towards more human-like emotion recognition by modelling the perception uncertainty},
  author={Han, Jing and Zhang, Zixing and Schmitt, Maximilian and Pantic, Maja and Schuller, Bj{\"o}rn},
  booktitle={Proceedings of the 25th ACM international conference on Multimedia},
  pages={890--897},
  year={2017}
}

@inproceedings{wu2022novel,
  title={A novel sequential monte carlo framework for predicting ambiguous emotion states},
  author={Wu, Jingyao and Dang, Ting and Sethu, Vidhyasaharan and Ambikairajah, Eliathamby},
  booktitle={ICASSP},
  pages={8567--8571},
  year={2022},
  organization={IEEE}
}

@article{wu2026shala,
  title={SHALA-LLM: Smartly Handling Ambiguous Labels in Aligning LLMs},
  author={Wu, Jingyao and Wang, Ashley and Ong, Keane and Liang, Paul Pu and Picard, Rosalind},
  journal={arXiv preprint arXiv:2606.05376},
  year={2026}
}

@inproceedings{wu2024can,
  title={Can Modelling Inter-Rater Ambiguity Lead To Noise-Robust Continuous Emotion Predictions?},
  author={Wu, Ya-Tse and Wu, Jingyao and Sethu, Vidhyasaharan and Lee, Chi-Chun},
  booktitle={Proc. Interspeech 2024},
  pages={3714--3718},
  year={2024}
}

@inproceedings{nie2020can,
  title={What can we learn from collective human opinions on natural language inference data?},
  author={Nie, Yixin and Zhou, Xiang and Bansal, Mohit},
  booktitle={Proceedings of the 2020 Conference on Empirical Methods in Natural Language Processing (EMNLP)},
  pages={9131--9143},
  year={2020}
}

@article{klie2024analyzing,
  title={Analyzing dataset annotation quality management in the wild},
  author={Klie, Jan-Christoph and Castilho, Richard Eckart de and Gurevych, Iryna},
  journal={Computational Linguistics},
  volume={50},
  number={3},
  pages={817--866},
  year={2024},
  publisher={MIT Press 255 Main Street, 9th Floor, Cambridge, Massachusetts 02142, USA~…}
}

@article{hullermeier2021aleatoric,
  title={Aleatoric and epistemic uncertainty in machine learning: An introduction to concepts and methods},
  author={H{\"u}llermeier, Eyke and Waegeman, Willem},
  journal={Machine learning},
  volume={110},
  number={3},
  pages={457--506},
  year={2021},
  publisher={Springer}
}
\end{document}